\PassOptionsToPackage{table}{xcolor}

\documentclass[10pt,twocolumn,letterpaper]{article}

\usepackage[pagenumbers]{wacv} 

\usepackage{graphicx}
\usepackage{amsmath}
\usepackage{amssymb}
\usepackage{booktabs}
\usepackage{pgfplots}
\usepackage{subcaption}
\usepackage{adjustbox}
\usepackage[table]{xcolor}

\definecolor{myLightGreen}{RGB}{222,254,209}

\usepackage[pagebackref,breaklinks,colorlinks]{hyperref}

\usepackage[capitalize]{cleveref}
\crefname{section}{Sec.}{Secs.}
\Crefname{section}{Section}{Sections}
\Crefname{table}{Table}{Tables}
\crefname{table}{Tab.}{Tabs.}

\def\wacvPaperID{2738} 
\def\confName{WACV}
\def\confYear{2025}

\begin{document}

\title{Point-GN: A Non-Parametric Network Using Gaussian Positional Encoding for Point Cloud Classification}

\author{Marzieh Mohammadi\thanks{These authors contributed equally to this work.} \hspace{1cm} Amir Salarpour\footnotemark[1]\\
Sirjan University of Technology, Iran\\
{\tt\small mrziehmohamadi.gmail.com, salarpour@sirjantech.ac.ir}
}

\maketitle

\begin{abstract}

This paper introduces \textbf{Point-GN}, a novel non-parametric network for efficient and accurate 3D point cloud classification. Unlike conventional deep learning models that rely on a large number of trainable parameters, Point-GN leverages non-learnable components—specifically, \textbf{Farthest Point Sampling (FPS)}, \textbf{k-Nearest Neighbors (k-NN)}, and \textbf{Gaussian Positional Encoding (GPE)}—to extract both local and global geometric features. This design eliminates the need for additional training while maintaining high performance, making Point-GN particularly suited for real-time, resource-constrained applications. We evaluate Point-GN on two benchmark datasets, \textbf{ModelNet40} and \textbf{ScanObjectNN}, achieving classification accuracies of \textbf{85.29\%} and \textbf{85.89\%}, respectively, while significantly reducing computational complexity. Point-GN outperforms existing non-parametric methods and matches the performance of fully trained models, all with zero learnable parameters. Our results demonstrate that Point-GN is a promising solution for 3D point cloud classification in practical, real-time environments. For more details, see the code at: \url{https://github.com/asalarpour/Point_GN}.

\end{abstract}


\section{Introduction}
\label{sec:intro}

Point cloud classification is a critical task in 3D data analysis and has been widely employed in various fields, including object detection \cite{zhangyu2021camera,zhou2018voxelnet}, 3D reconstruction \cite{xu2021review}, robotics \cite{varley2017shape}, and medicine \cite{guo2018deep,yu20213d}. Unlike 2D images that are structured in regular grids, point clouds consist of unordered and irregular sets of points, presenting unique challenges for efficient and accurate analysis. The unordered nature and high dimensionality of point clouds make traditional 2D image processing techniques unsuitable for this task, thereby necessitating specialized algorithms that can handle the unique structure of 3D data.

\begin{figure} [t]
    \centering
    \includegraphics[width=0.65\textwidth, trim=140 640 30 0, clip]{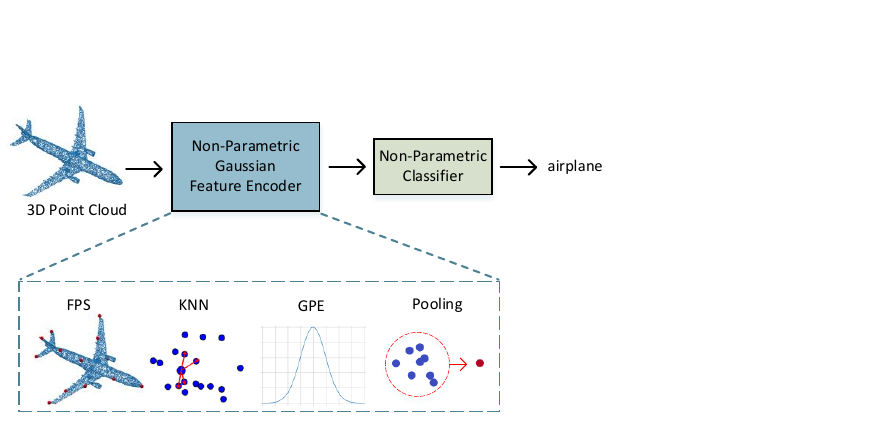}
    \caption{Illustration of the proposed non-parametric network for point cloud classification.}
   \label{fig:illustration}
\end{figure}

Deep learning has significantly advanced point cloud classification, with models like PointNet \cite{qi2017pointnet} and PointNet++ \cite{qi2017pointnet++} enabling direct processing of point clouds without requiring 3D voxelization or mesh generation. By treating point clouds as unordered sets of points, these models learn permutation-invariant features. However, their reliance on large numbers of parameters leads to higher memory usage and longer training times. For example, PointNet++ extends PointNet with hierarchical structures and local feature capture but also increases computational complexity, which can hinder scalability and real-time performance.

The trend toward more complex models further compounds the issue. For instance, CurveNet \cite{muzahid2020curvenet}, which focuses on curve-based feature extraction, increases training times by up to 10 times compared to PointNet++, with little improvement in performance. Similarly, PointMLP \cite{ma2022rethinking} adds 11.9 million parameters over PointNet++, achieving only a modest 0.5\% gain in precision. This growing complexity underscores the need for more efficient models that balance accuracy and computational cost, especially for real-time applications like autonomous driving \cite{qi2018frustum}, AR/VR, and medical diagnostics.

To address the high computational and memory demands of parametric models, recent research has explored non-parametric methods. These approaches, such as PointClip \cite{zhang2022pointclip}, which uses pre-trained 2D models for point cloud classification, and Point-NN \cite{zhang2023parameter}, which avoids learnable weights, aim to reduce parameter sets and improve efficiency. However, these methods often trade off computational efficiency for classification accuracy, highlighting the need for further improvement.

In this paper, we propose a novel non-parametric method for point cloud classification that introduces a Gaussian function for positional embedding, as shown in \cref{fig:illustration}. This Gaussian embedding enhances classification accuracy by capturing local geometric structures without requiring extra parameters or retraining. Integrated into the non-parametric framework of PointNet++ with Farthest Point Sampling (FPS) and k-Nearest Neighbors (k-NN), it improves performance while maintaining computational efficiency.

Our contributions are threefold:
\begin{itemize}
    \item We introduce a \textbf{Gaussian embedding function} to the non-parametric framework, which significantly improves classification accuracy without adding computational overhead.
    \item We \textbf{simplify the network design}, eliminating unnecessary complexities while maintaining performance comparable to state-of-the-art parametric models, offering a more efficient alternative for resource-constrained environments.
    \item We demonstrate the \textbf{efficiency and scalability} of our approach through extensive experiments on popular benchmarks, including ModelNet40 \cite{wu20153d} and ScanObjectNN \cite{uy2019revisiting}, achieving competitive performance without additional costs in terms of memory or computational requirements.
\end{itemize}

This method offers a practical solution for efficient, accurate point cloud classification, ideal for real-time applications with limited resources, such as robotics and autonomous systems \cite{aldeeniv, mohajeransari2024discovering}, where both accuracy and efficiency are essential.

\section{Related works}
\label{sec:related}
In this section, we provide an overview of prior approaches to 3D point cloud classification, focusing on both projection-based and point-based methods, and introduce recent advancements in positional encoding techniques, which have significantly impacted model performance and efficiency.

\subsection{3D Point Cloud Classification}
3D point cloud classification methods can be divided into projection-based and point-based approaches.

Projection-based methods convert 3D point clouds into 2D representations, such as depth maps \cite{feng2018gvcnn,su2015multi,goyal2021revisiting} or voxel grids \cite{maturana2015voxnet,liu2019point,wang2018msnet}, enabling the use of 2D image processing techniques. However, they often lose spatial details due to the sparsity and incompleteness of point clouds.

Point-based methods process raw 3D point clouds directly, preserving geometric information. PointNet \cite{qi2017pointnet} processes points independently and aggregates global features using max pooling but struggles with local geometric details. PointNet++ \cite{qi2017pointnet++} improves this by introducing a hierarchical architecture for capturing local features. Other advancements include convolutional \cite{li2018pointcnn,thomas2019kpconv,peyghambarzadeh2020point, salarpour2014long} and graph-based models \cite{wang2019dynamic, lu2020pointngcnn}, as well as attention and transformer-based methods \cite{zhao2021point, yang2019modeling, yu2022point} that model long-range dependencies.



However, these models are computationally intensive, limiting their use in real-time applications. Efficient methods like ShellNet \cite{zhang2019shellnet} and RandLA-Net \cite{hu2020randla} reduce memory usage, but still face challenges with large-scale data. Convolutional models like KPConv \cite{thomas2019kpconv} reduce memory overhead through sparse convolutions but still require significant resources.


Inspired by Point-NN \cite{zhang2023parameter}, we propose a novel non-parametric model for point cloud classification that improves feature extraction without introducing additional trainable parameters, thus enhancing both efficiency and scalability.

\subsection{Positional Encodings}
Positional encoding was first introduced in the Transformer architecture \cite{vaswani2017attention} to inject positional information into input sequences, such as words in a sentence, using sinusoidal functions. This method has since been widely adopted in natural language processing and computer vision, where capturing spatial relationships is critical. In point cloud processing, positional encoding allows the model to retain spatial awareness in 3D space, particularly for applications involving coordinate-based models, such as 2D image synthesis \cite{nguyen2015deep} or 3D scene reconstruction \cite{niemeyer2020differentiable}.

In 3D point cloud processing, positional encoding plays a crucial role in capturing the underlying geometric relationships between points. One notable application is in Neural Radiance Fields (NeRF) \cite{mildenhall2021nerf}, where sinusoidal encoding transforms input coordinates into higher-dimensional feature spaces, enabling the accurate reconstruction of fine-grained details in 3D scenes. Such approaches demonstrate the importance of encoding spatial information when dealing with high-frequency signals, as it accelerates network convergence and enhances the model's ability to capture complex geometric structures.

Building on these advancements, we propose a novel positional encoding scheme based on Gaussian functions, specifically tailored for 3D point cloud classification. Unlike sinusoidal encodings, Gaussian embeddings allow for more flexible representation of spatial relationships in 3D space, offering improved feature extraction and classification performance. Our approach demonstrates that Gaussian embeddings can match, and in some cases surpass, the performance of traditional sinusoidal encodings, particularly when applied to point clouds in non-parametric frameworks.

\section{The Proposed Method}
\label{sec:method}
\begin{figure}[t]
  \centering
   \includegraphics[width=0.5\textwidth, trim=22 430 30 20, clip]{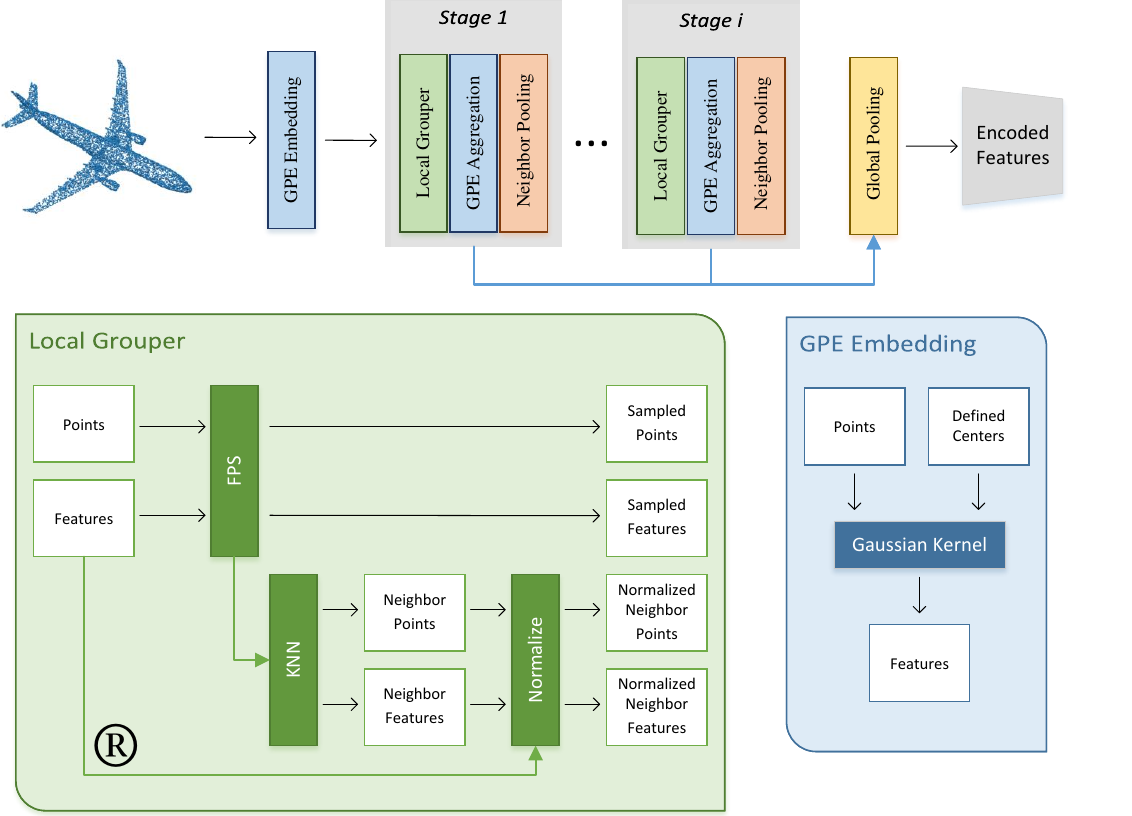}
    \caption{Overview of the architecture of our Non-Parametric Feature Encoder. The figure illustrates the main components of Point-GN, including Gaussian Positional Encoding (GPE), local grouping and feature aggregation. Each stage of the network is designed to efficiently capture spatial relationships within the point cloud without the need for learnable parameters.}

  \label{fig:framework}
\end{figure}

In this section, we present the details of \textbf{Point-GN}, our Non-Parametric Network that utilizes Gaussian Positional Encoding (GPE) for point cloud classification. \cref{fig:framework} provides an overview of the Point-GN framework, which highlights the key stages in processing point cloud data. To ground the discussion, we first revisit the fundamental concepts behind 3D point clouds and how they are typically classified. We then elaborate on the design of our non-parametric feature encoder and classifier, which are integral components of Point-GN.

\subsection{Background}

A 3D point cloud is a collection of points in 3D space representing the shape or structure of an object or scene. Each point $\mathbf{p}_i = (x_i, y_i, z_i) \in \mathbb{R}^3$ is defined by its coordinates and may have additional attributes, such as color or surface normal vectors. Given a point cloud $\mathcal{P} = \{\mathbf{p}_1, \mathbf{p}_2, \dots, \mathbf{p}_N\}$, where $N$ is the number of points, each point $\mathbf{p}_i$ is represented by its coordinates $(x_i, y_i, z_i)$. An encoder extracts meaningful information from the point cloud into a compact representation:

\begin{equation}
\text{Encoder}(\mathcal{P}) = \mathbf{F} \in \mathbb{R}^d 
\end{equation}

where $\mathbf{F}$ is the feature vector that encodes the essential characteristics of the point cloud into a $d$-dimensional space.

For classification, the feature vector $\mathbf{F}$ is fed into a classifier that maps the feature vector to $C$ classes, producing a vector of logits $\mathbf{y} \in \mathbb{R}^C$:

\begin{equation}
\text{Classifier}(\mathbf{F}) = \mathbf{y} = (y_1, y_2, \dots, y_C)
\end{equation}

The predicted class $c$ is determined by selecting the class with the highest score:

\begin{equation}
c = \arg\max_i y_i
\end{equation}

\subsection{Gaussian Positional Encoding (GPE)}

Gaussian Positional Encoding (GPE) embeds spatial information into the feature representation of individual points in the 3D point cloud. By transforming raw point coordinates into a higher-dimensional space, GPE provides the model with richer spatial context without introducing learnable parameters. The encoding is formulated as:

\begin{equation}
\gamma_x(x_i, v_j) = \exp\left(-\frac{\|x_i - v_j\|^2}{2\sigma^2}\right)
\end{equation}
\begin{equation}
\gamma_y(y_i, v_j) = \exp\left(-\frac{\|y_i - v_j\|^2}{2\sigma^2}\right)
\end{equation}
\begin{equation}
\gamma_z(z_i, v_j) = \exp\left(-\frac{\|z_i - v_j\|^2}{2\sigma^2}\right)
\end{equation}

where $v_j$ are predefined reference points, and $\sigma$ is the standard deviation that controls the focus on local vs. global spatial information. A smaller $\sigma$ captures local detail, while a larger $\sigma$ captures broader spatial patterns.

The encoded feature vector for each point is:

\begin{equation}
\gamma(\mathbf{p}_i) = \left[\gamma_x(x_i, v_j), \gamma_y(y_i, v_j), \gamma_z(z_i, v_j)\right]_{j=1}^V
\end{equation}

where $V$ is the number of reference points along each axis.

\subsection{Non-Parametric Feature Encoder}

In our approach, the non-parametric feature encoder leverages Gaussian Positional Encoding (GPE) to capture and aggregate spatial information from 3D point clouds. This hierarchical encoder adapts to various input configurations without relying heavily on learned parameters, making it flexible across different tasks.

\subsubsection{GPE Embedding}

The embedding process begins by applying GPE to each point $\mathbf{p}_i$ in the point cloud $\mathcal{P}$. This transformation maps the raw 3D coordinates into a higher-dimensional feature space, enhancing the model’s ability to understand spatial relationships. As a result, each point $\mathbf{p}_i$ is represented by a richer feature vector $\gamma(\mathbf{p}_i)$, capturing its spatial relationships more effectively.

This transformation expands the original 3D coordinates into a $V \times 3$-dimensional space, where $V$ represents the number of reference points $v_j$ used in the encoding. These reference values are strategically chosen or learned by the model, often distributed uniformly between $-1$ and $1$. The parameter $\sigma$, controlling the width of the Gaussian function, determines how spatial information is captured. A smaller $\sigma$ focuses on local details, while a larger $\sigma$ captures more global spatial relationships. This flexibility enables the GPE to balance the capture of both local and global spatial information effectively.

The GPE embedding serves as the foundational step in our non-parametric feature encoder. By transforming raw 3D point clouds into a feature space rich in spatial context, $\gamma(\mathbf{p}_i)$ enhances the model's ability to recognize and utilize the underlying geometric structures of the data. This process is integral to the effectiveness of the encoder, leading to improved performance in tasks such as object recognition, segmentation, and classification, where a deep understanding of both local and global spatial relationships is essential.

\subsubsection{Local Grouper}

After GPE embedding, feature extraction proceeds through multiple stages, each involving a local grouper, GPE aggregation, and neighbor pooling. At each stage, the input point cloud from the previous stage is represented as $\{\mathbf{p}_i, \gamma(\mathbf{p}_i)\}_{i=1}^N$, where $\mathbf{p}_i \in \mathbb{R}^3$ denotes the coordinates of point $i$ and $\gamma(\mathbf{p}_i) \in \mathbb{R}^{V \times 3}$ represents the GPE-embedded features of that point.

The process begins with Farthest Point Sampling (FPS) to downsample the number of points from $N$ to $N/2$, selecting a subset of local center points:

\begin{equation}
\{\mathbf{p}_j, \gamma(\mathbf{p}_j)\}_{j=1}^{N/2} = \text{FPS}\left(\{\mathbf{p}_i, \gamma(\mathbf{p}_i)\}_{i=1}^N\right)
\end{equation}

Next, the downsampled point coordinates $\mathbf{p}_j$ and the original point coordinates $\mathbf{p}_i$ are used by the K-Nearest Neighbors (KNN) algorithm to find the $K$ nearest neighbors for each downsampled point $\mathbf{p}_j$. The indices of these nearest neighbors are used to retrieve the corresponding coordinates and features:

\begin{equation}
\text{idx}_j = \text{KNN}(\mathbf{p}_j, \mathbf{p}_i)
\end{equation}
The retrieved coordinates and features are:
\begin{equation}
\mathbf{P}_j = \text{retrieve}\left(\{\mathbf{p}_i\}_{i=1}^N ,\text{idx}_j\right) \in \mathbb{R}^{K \times 3}
\end{equation}

\begin{equation}
\Gamma_j = \text{retrieve}\left(\{\gamma(\mathbf{p}_i)\}_{i=1}^N ,\text{idx}_j\right) \in \mathbb{R}^{K \times (V \times 3)}
\end{equation}

Here, $\mathbf{P}_j$ represents the gathered coordinates, and $\Gamma_j$ represents the gathered features for the point $\mathbf{p}_j$. The retrieved coordinates $\mathbf{P}_j$ and features $\Gamma_j$ are then normalized using the mean and standard deviation of each point’s neighbors. These normalized coordinates and features are then passed to the next stage for further processing.

\subsubsection{GPE Aggregation}
The features from the Local Grouper are then fed into the GPE Aggregation module. Here, GPE is applied to the retrieved coordinates $\mathbf{P}_j$ to encode spatial information. These encoded features are then weighted and combined with retrieved features $\Gamma_j$, emphasizing points closer to the center. The updated feature representation is:

\begin{equation}
\Gamma_j \leftarrow \Gamma_j + \gamma(\mathbf{P}_j) \odot \gamma(\mathbf{P}_j)
\end{equation}

In this formulation, $\gamma(\mathbf{P}_j)$ represents the encoded spatial information of the retrieved neighbors, and $\Gamma_j$ represents the features retrieved from the nearest neighbors. The element-wise multiplication $\odot$ ensures that the final aggregated features are influenced by both local features and spatial encoding, effectively capturing detailed local geometry while preserving spatial relationships.

\subsubsection{Neighbor Pooling}

Following GPE Aggregation, the neighbor pooling process aggregates features using both mean and max operations across the neighbor dimension. For each point, the pooled features are calculated as:

\begin{equation}
\Phi_j = \text{Mean}(\Gamma_j) + \text{Max}(\Gamma_j), \quad \forall j \in \{1, \ldots, N/2\}
\end{equation}

Here, Mean($\Gamma_j$) and Max($\Gamma_j$) are permutation-invariant operations, ensuring that the order of neighbors does not affect the pooled features.

\subsubsection{Aggregation Across Stages}

The non-parametric feature encoder includes four stages, each producing pooled features $\Phi_j^s$. After processing through all stages, global pooling is applied to the results from each stage. The final feature vector $\mathbf{F}$ for the non-parametric feature encoder is obtained by concatenating the global mean and max features from all four stages:

\begin{equation} 
\mathbf{F} = \bigoplus_{s=1}^4 \left[\text{Mean}(\Phi_j^s) + \text{Max}(\Phi_j^s)\right] 
\end{equation}

This formulation captures and aggregates spatial and feature information across multiple levels by combining the mean and max features from each stage.

\subsection{Non-Parametric Classifier}

To preserve the non-parametric nature of Point-GN, we adopt a similarity-based classification approach \cite{zhang2023parameter}. Given a test point cloud, its feature vector $\mathbf{F}_{\text{test}}$ is compared to the feature embeddings $\mathbf{F}_{\text{train}}$ from the training set. The classifier computes a similarity score between the test and training embeddings, which is then used to assign a class label based on the closest matching features. This similarity-driven mechanism avoids the need for traditional parametric models, maintaining the flexibility and efficiency of the non-parametric framework.

\begin{equation}
\mathbf{F}_{\text{train}} = \bigoplus_{m=1}^{M} \mathbf{F}_m
\end{equation}

\begin{equation}
\mathbf{L}_{\text{train}} = \bigoplus_{m=1}^{M} \mathbf{L}_m
\end{equation}

\begin{figure}
  \centering
    \includegraphics[width=0.57\textwidth, trim=120 550 10 70, clip]{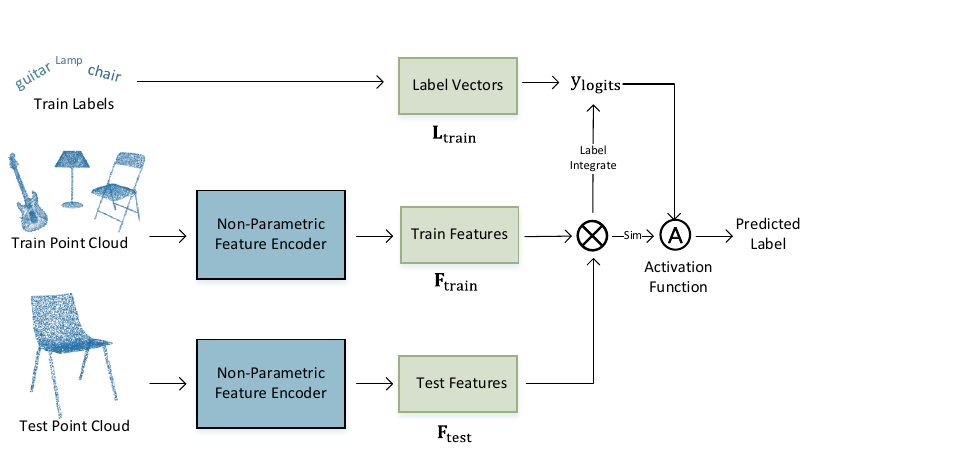}
    \vspace{-5mm}
    \caption{Illustration of the non-parametric classifier pipeline. The test feature is compared with the stored feature embeddings from the training set, and similarity scores are computed to assign the most likely class label based on proximity in feature space.}
  \label{fig:classifier}
\end{figure}

\subsubsection{Feature Representation and Label Embedding}

Given a training set of $M$ point clouds $\{\mathcal{P}_m\}_{m=1}^M$, each belonging to one of $C$ categories, we first extract a global feature vector $\mathbf{F}_m$ for each point cloud $\mathcal{P}_m$ using the non-parametric encoder. The corresponding labels $\{y_m\}_{m=1}^M$ are transformed into embedded label vectors $\mathbf{L}_m$.

The process of storing these feature and label embeddings is shown in \cref{fig:classifier}. The feature embeddings are stored in the global feature matrix $\mathbf{F}_{\text{train}}$, and the label embeddings are stored in the label matrix $\mathbf{L}_{\text{train}}$, both defined as follows:

\subsubsection{Similarity-Based Classification}

For a test point cloud $\mathcal{P}_{\text{test}}$, the non-parametric encoder generates the feature vector $\mathbf{F}_{\text{test}}$. We compute the similarity between $\mathbf{F}_{\text{test}}$ and the stored training features in $\mathbf{F}_{\text{train}}$ using the following equation:

\begin{equation}
\text{Sim} = \mathbf{F}_{\text{test}} \cdot \mathbf{F}_{\text{train}}^T
\end{equation}

The similarity scores in $\text{Sim}$ are used to weight the corresponding label embeddings from $\mathbf{L}_{\text{train}}$. The final classification logits are computed using the following activation function:

\begin{equation}
\mathbf{y}_{\text{logits}} = \exp\left(-\gamma \cdot (1 - \text{Sim})\right) \cdot \mathbf{L}_{\text{train}}
\end{equation}

Here, $\gamma$ is a scaling factor, and $\exp(-\gamma \cdot (1 - \text{Sim}))$ serves as the activation function adapted from Tip-Adapter \cite{zhang2021tip}, where higher similarity scores result in stronger contributions from the corresponding labels in $\mathbf{L}_{\text{train}}$.

\subsubsection{Classification Decision}

The predicted class label is determined by applying an activation function to the logits, selecting the category with the highest value. In our case, we use a softmax activation for this final step:

\begin{equation}
c = \arg\max(\text{softmax}(\mathbf{y}_{\text{logits}}))
\end{equation}

Through this similarity-based label integration, the classifier is able to effectively differentiate between various point cloud instances using a simple and efficient mechanism.

\section{Experiments}
\label{sec:exper}
\begin{table}
  \centering
  {\small{
  \begin{tabular}{@{}lcc@{}}
    \toprule
    Method & Acc. (\%) & Param. \\
    \midrule
    PointNet \cite{qi2017pointnet} & 89.2 & 3.5 M \\
    PointNet++ \cite{qi2017pointnet++} & 90.7 & 1.5 M \\
    PointCNN \cite{li2018pointcnn} & 92.2 & 0.6 M \\
    DGCNN \cite{wang2019dynamic} & 92.9 & 1.8 M \\
    GBNet \cite{qiu2021geometric} & 93.8 & 8.4 M \\
    CurveNet \cite{xiang2021walk} & 93.8 & 2.0 M \\
    PointNext-S \cite{qian2022pointnext} & 93.2 & 1.4 M \\
    PointMLP \cite{ma2022rethinking} & \textbf{94.1} & 12.6 M \\
    \midrule
    Point-NN \cite{zhang2023parameter} & 81.8 & \textbf{0.0 M}  \\ 
    \rowcolor{myLightGreen}
    Point-GN (ours) &  \textbf{85.3} & \textbf{0.0 M}  \\
    \bottomrule
  \end{tabular}
  }}
  \caption{\textbf{Shape Classification on Synthetic ModelNet40 \cite{wu20153d}.} All compared methods take 1,024 points as input. We report the accuracy without the voting strategy.}
  \label{tab:modelnet40}
\end{table}

In this section, we benchmark the performance of \textbf{Point-GN} against state-of-the-art methods for 3D shape classification. We conduct experiments on two widely recognized datasets: \textbf{ModelNet40} \cite{wu20153d} and \textbf{ScanObjectNN} \cite{uy2019revisiting}. These datasets were selected for their complementary characteristics: ModelNet40 comprises clean, synthetic 3D models, while ScanObjectNN presents more challenging real-world scenarios with occlusions and background noise. By evaluating on these datasets, we aim to demonstrate the robustness and versatility of our approach across both synthetic and real-world data. Additionally, we compare Point-GN to existing non-parametric methods to showcase its efficiency and competitiveness.

\subsection{Experimental Setup}

We evaluate the performance of our \textbf{Point-GN} method on a system equipped with an NVIDIA RTX 4090 GPU. Although \textbf{Point-GN} is a non-parametric method and does not require traditional model training, the GPU significantly accelerates the inference process, allowing for efficient evaluation across the large and diverse \textbf{ModelNet40} \cite{wu20153d} and \textbf{ScanObjectNN} \cite{uy2019revisiting} datasets. This high-performance hardware ensures that our approach can handle the complexity and size of real-world 3D data, providing rapid evaluations during the benchmarking process.

\subsection{Dataset Details}

The \textbf{ModelNet40} \cite{wu20153d} dataset consists of 12,311 CAD models across 40 object categories, split into 9,843 samples for training and 2,468 for testing. This dataset is widely used for point cloud classification due to its clean, synthetic nature, providing a controlled environment for benchmarking.

In contrast, the \textbf{ScanObjectNN} \cite{uy2019revisiting} dataset presents a more challenging real-world scenario, with 2,902 samples across 15 object categories. Objects in ScanObjectNN are often occluded, cluttered, or contain background noise, providing a closer simulation to real-world 3D data. The dataset is divided into three official subsets: \textbf{OBJ-BG}, which contains objects with background noise, \textbf{OBJ-ONLY}, with objects without background, and \textbf{PB-T50-RS}, featuring partial occlusions and transformations. These subsets test the robustness of models under various degrees of complexity.

For both datasets, we follow the common practice of sampling 1,024 points from each object, as used in prior works (e.g., PointNet++ \cite{qi2017pointnet++}, DGCNN \cite{wang2019dynamic}). Our model combines \textbf{maximum pooling} and \textbf{average pooling} to enhance feature aggregation, inspired by DGCNN \cite{wang2019dynamic}.

\begin{table}
  \centering
   {\small{
    \resizebox{0.47\textwidth}{!}{ 
  \begin{tabular}{@{}l@{\hspace{0pt}}c@{\hspace{3pt}}c@{\hspace{3pt}}c@{\hspace{3pt}}c@{}}
    \toprule
    Method  &OBJ-BG & OBJ-ONLY & PB-T50-RS & Param. \\
    \midrule
    3DmFV \cite{ben20183dmfv} &\textcolor{blue}{68.2} & \textcolor{blue}{73.8} & \textcolor{blue}{63.0} & - \\
    PointNet \cite{qi2017pointnet} &\textcolor{blue}{73.3} & \textcolor{blue}{79.2} & \textcolor{blue}{68.2} & 3.5 M\\
    PointNet++ \cite{qi2017pointnet++} &\textcolor{blue}{82.3} &\textcolor{blue}{84.3} & \textcolor{blue}{77.9} & 1.5 M \\
    DGCNN \cite{wang2019dynamic} & \textcolor{blue}{82.8} &\textbf{86.2} &\textcolor{blue}{78.1} & 1.8 M \\
    PointCNN \cite{li2018pointcnn} &\textbf{86.1} & \textcolor{blue}{85.5}& \textcolor{blue}{78.5} & -\\
    GBNet \cite{qiu2021geometric} &- & -& \textcolor{blue}{80.5} &8.4 M \\
    PointMLP \cite{ma2022rethinking}  & -&- & \textcolor{blue}{85.4} & 12.6 M\\
    PointNeXt-S \cite{qian2022pointnext}  & -&- & {87.7} & 1.5 M\\
    PointMetaBase-S \cite{lin2023meta}  & -&- & \textbf{87.9} & 0.6 M\\
    \midrule
    Point-NN \cite{zhang2023parameter} & {71.1}&{74.9} & {64.9} & 0.0 M\\ 
    \rowcolor{myLightGreen}
    Point-GN (ours) &  \textbf{85.2} & \textbf{86.0} & \textbf{86.4} & \textbf{0.0 M}\\
    
    \bottomrule
    
  \end{tabular} 
  }}}
  \caption{\textbf{Shape Classification on the Real-world ScanObjectNN \cite{uy2019revisiting}.} We report the accuracy (\%) on three official splits of ScanObjectNN: OBJ-BG, OBJ-ONLY, and PB-T50-RS. The results in blue correspond to fully trained models, which show lower accuracy than our train-free method, Point-GN, which outperforms all others.}
  \label{tab:scanobjectnn1}
\end{table}

\subsection{Shape Classification on ModelNet40}

We evaluate the performance of Point-GN on the \textbf{ModelNet40} \cite{wu20153d} dataset in \cref{tab:modelnet40}. Point-GN achieves an accuracy of \textbf{85.3\%}, demonstrating strong performance in synthetic 3D shape classification. This result highlights Point-GN's ability to effectively capture both local and global geometric features, all while maintaining a minimal model complexity.

When compared to the non-parametric Point-NN \cite{zhang2023parameter}, Point-GN shows a \textbf{+3.5\%} improvement in accuracy, despite having zero trainable parameters. This demonstrates the effectiveness of our approach in extracting meaningful features without relying on large parameter counts. Furthermore, Point-GN achieves an inference speed of \textbf{301 samples/second} (measured on our system), ensuring high efficiency for real-time applications. This is especially notable when compared to parametric models like PointMLP \cite{ma2022rethinking}, which requires 12.6M parameters to achieve a slightly higher accuracy of 94.1\%.

The combination of competitive accuracy and exceptional computational efficiency makes Point-GN an attractive choice for resource-constrained environments, where real-time performance and minimal model complexity are crucial.

\begin{table}
    \centering
    \begin{tabular}{@{}lccc@{\hspace{6pt}}c@{}}

        \toprule
        {Method} & \multicolumn{2}{c}{{5-way}} & \multicolumn{2}{c}{{10-way}} \\
        \cmidrule(lr){2-3} \cmidrule(lr){4-5}
        & {10-shot} & {20-shot} & {10-shot} & {20-shot} \\
        \midrule
        DGCNN \cite{wang2019dynamic}& 31.6 & 40.8 & 19.9 & 16.9 \\
        FoldingNet \cite{yang2018foldingnet} & 33.4 & 35.8 & 18.6 & 15.4 \\
        PointNet++ \cite{qi2017pointnet++} & 38.5 & 42.4 & 23.0 & 18.8 \\
        PointNet \cite{qi2017pointnet} & 52.0 & 57.8 & 46.6 & 35.2 \\
        3D-GAN \cite{wu2016learning}& 55.8 & 65.8 & 40.3 & 48.4 \\
        PointCNN \cite{li2018pointcnn}& 65.4 & 68.6 & 46.6 & 50.0 \\
        \midrule
        Point-NN \cite{zhang2023parameter}& 88.8 & \textbf{90.9}  & 79.9 & 84.9 \\
        \rowcolor{myLightGreen}
        Point-GN (ours) & \textbf{90.7} & \textbf{90.9} & \textbf{81.6} & \textbf{86.4} \\
        \bottomrule
    \end{tabular}
    \caption{\textbf{Few-shot Classification on ModelNet40 \cite{wu20153d}.} We compute the mean accuracy (\%) across 10 separate runs. The presented results of existing methods are sourced from \cite{sharma2020self}.}
    \label{tab:fewshot}
\end{table}

\subsection{Shape Classification on ScanObjectNN}

On the \textbf{ScanObjectNN} \cite{uy2019revisiting} dataset (\cref{tab:scanobjectnn1}), Point-GN demonstrates superior performance in real-world scenarios, outperforming most existing methods across all three official splits: OBJ-BG, OBJ-ONLY, and PB-T50-RS. Notably, \textbf{Point-NN} \cite{zhang2023parameter}, the only other non-parametric method for comparison, is significantly outperformed by Point-GN. In the most challenging split, \textbf{PB-T50-RS}, Point-GN achieves a \textbf{+21.5\%} improvement in accuracy over Point-NN, highlighting its robustness to occlusions and background noise in real-world data.

Rather than using different setups for each split, we adopted a single configuration for all three splits and aimed to find the best average performance. The \textbf{average accuracy} across the three splits for Point-GN is \textbf{85.89\%}, demonstrating its consistency across varying conditions. This approach ensures we identify a setup that works best on average, rather than optimizing separately for each split.

When compared to fully trained models (shown in blue in \cref{tab:scanobjectnn1}), Point-GN consistently outperforms most, achieving higher accuracy across all splits. In particular, it surpasses models like PointNet \cite{qi2017pointnet}, PointNet++ \cite{qi2017pointnet++}, and PointMLP \cite{ma2022rethinking} by substantial margins. The difference between Point-GN and the top-performing model, PointMetaBase-S \cite{lin2023meta}, is less than 2\%, indicating that Point-GN competes closely with state-of-the-art methods despite having zero trainable parameters.

These results underscore the power of Point-GN's non-parametric design, which offers competitive performance while maintaining computational efficiency and avoiding the complexity of parameter-heavy models.

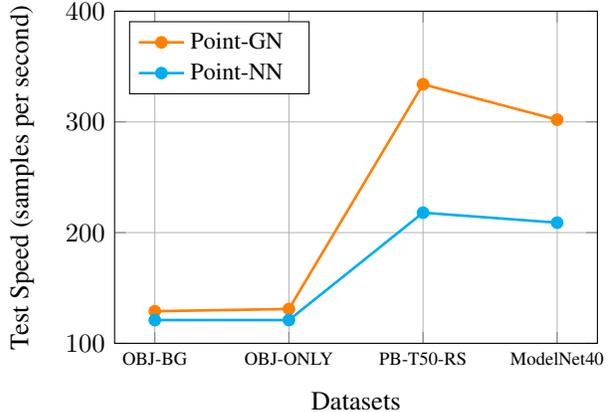
\begin{figure}[htbp]
    \begin{center}
    \begin{tikzpicture}
        \begin{axis}[
            width=8cm, height=6cm,
            xlabel={Datasets},
            ylabel={Test Speed (samples per second)},
            title={},
            xtick={1, 2, 3, 4},
            xticklabels={OBJ-BG, OBJ-ONLY, PB-T50-RS, ModelNet40},
            xticklabel style={font=\scriptsize},
            grid=both,
            grid style={line width=.1pt, draw=gray!30},
            major grid style={line width=.2pt, draw=gray!60},
            legend style={font=\small, text width=1.4cm, align=left},
            legend pos=north west,
            ymin=100, ymax=400 
        ]
            \addplot[
                color=orange,
                mark=*, 
                mark options={fill=orange, draw=orange, thick}, 
                line width=1pt
            ] coordinates {
                (1, 129) (2, 131) (3, 334) (4, 302)
            };
            \addlegendentry{Point-GN};
    
            \addplot[
                color=cyan,
                mark=*, 
                mark options={fill=cyan, draw=cyan, thick},
                line width=1pt
            ] coordinates {
                (1, 121) (2, 121) (3, 218) (4, 209)
            };
            \addlegendentry{Point-NN};
        \end{axis}
    \end{tikzpicture}
    \end{center}
    \vspace{-5mm}
    \caption{\textbf{Test Speed (samples per second) on ScanObjectNN \cite{uy2019revisiting} and ModelNet40 \cite{wu20153d} datasets.} The plot compares the inference speed of Point-NN \cite{zhang2023parameter} and Point-GN on four different datasets. Point-GN shows significant improvements in inference speed across all datasets.}
    \label{fig:test_speed}
\end{figure}

\begin{figure*}[htbp]
    \centering
    \begin{minipage}[t]{0.3\textwidth}
        \centering
        \begin{tikzpicture}
            \begin{axis}[
            width=\textwidth,
            height=5cm,
            xmin=70, xmax=130,
            xtick={70,80,90,100,110,120,130},
            xticklabel style={font=\scriptsize},
            xlabel={(a) K},
            xlabel style={font=\small, yshift=1.5ex},
            ymin=80, ymax=88,
            ylabel={Accuracy (\%)},
            ylabel style={font=\small, yshift=-2.5ex},
            yticklabel style={font=\scriptsize},
            grid=major,
            legend style={at={(2,1.1)}, anchor=south, legend columns=4, font=\scriptsize},
            ]

            \addplot[color=orange, thick, dashed] coordinates {(70, 83.13) (80,83.51) (90,83.61) (100,83.85) (110,84.03) (120,84.18) (130,83.89)};
            \addlegendentry{ModelNet40 Average}
            \addplot[color=orange, thick] coordinates {(70, 84.44) (80,84.76) (90,84.68) (100,85.09) (110,85.13) (120,85.29) (130,84.97)};
            \addlegendentry{ModelNet40 Best}
            \addplot[color=cyan, thick, dashed] coordinates {(70, 81.13) (80,81.47) (90,81.70) (100,81.83) (110,81.91) (120,82.09) (130,81.25)};
            \addlegendentry{ScanObjectNN Average}
            \addplot[color=cyan, thick] coordinates {(70, 85.40) (80,85.54) (90,85.48) (100,85.74) (110,85.63) (120,85.90) (130,85.08)};
            \addlegendentry{ScanObjectNN Best}
            
            \end{axis}
        \end{tikzpicture}
    \end{minipage}%
    \hspace{-0.75cm}
    \begin{minipage}[t]{0.4\textwidth}
        \centering
        \begin{tikzpicture}
            \begin{axis}[
            width=\textwidth,
            height=5cm,
            xmin=18, xmax=99,
            xtick={18,27,36,45,54,63,72,81,90,99},
            xticklabel style={font=\scriptsize},
            xlabel={(b) Dimension},
            xlabel style={font=\small, yshift=1.5ex},
            ymin=80, ymax=88,
            yticklabels={},
            grid=major,             
            ]
            \addplot[color=orange, thick, dashed] coordinates {(18,83.38) (27,83.77) (36,83.77) (45,83.79) (54,83.78) (63,83.78) (72,83.78) (81,83.77) (90,83.78) (99,83.77)};
            \addplot[color=orange, thick] coordinates {(18,85.09) (27,85.29) (36,85.29) (45,85.13) (54,85.13) (63,85.25) (72,85.13) (81,85.17) (90,85.17) (99,85.17)};
            \addplot[color=cyan, thick, dashed] coordinates {(18,82.06) (27,81.77) (36,81.71) (45,81.73) (54,81.60) (63,81.62) (72,81.52) (81,81.52) (90,81.46) (99,81.45)};
            \addplot[color=cyan, thick] coordinates {(18,85.29) (27,85.90) (36,85.45) (45,85.76) (54,85.67) (63,85.55) (72,85.60) (81,85.67) (90,85.67) (99,85.59)};
            \end{axis}
        \end{tikzpicture}
    \end{minipage}%
    \hspace{-0.75cm}
    \begin{minipage}[t]{0.14\textwidth}
        \centering
        \begin{tikzpicture}
            \begin{axis}[
            width=\textwidth,
            height=5cm,
            xmin=2, xmax=4,
            xtick={2, 3, 4},
            xticklabel style={font=\scriptsize},
            xlabel={(c) Stage},
            xlabel style={font=\small, yshift=1.5ex},
            ymin=80, ymax=88,
            yticklabels={},
            grid=major,
            ]
            \addplot[color=orange, thick, dashed] coordinates {(2,83.05) (3,83.96) (4,84.28)};
            \addplot[color=orange, thick] coordinates {(2,84.28) (3,85.09) (4,85.29)};
            \addplot[color=cyan, thick, dashed] coordinates {(2,79.20) (3,82.25) (4,83.79)};
            \addplot[color=cyan, thick] coordinates {(2,82.67) (3,85.08) (4,85.90)};
            \end{axis}
        \end{tikzpicture}
    \end{minipage}%
    \hspace{-0.75cm}  
    \begin{minipage}[t]{0.25\textwidth}
        \centering
        \begin{tikzpicture}
            \begin{axis}[
            width=\textwidth,
            height=5cm,
            xmin=0.2, xmax=0.45,
            xtick={0.2, 0.25, 0.3, 0.35, 0.4, 0.45},
            xticklabel style={font=\scriptsize},
            xlabel={(d) Sigma},
            xlabel style={font=\small, yshift=1.5ex},
            ymin=80, ymax=88,
            yticklabels={},
            grid=major,
            ]
            \addplot[color=orange, thick, dashed] coordinates {(0.2,82.71) (0.25,83.53) (0.3,83.83) (0.35,84.08) (0.4,84.19) (0.45,84.09)};
            \addplot[color=orange, thick] coordinates {(0.2,84.04) (0.25,84.68) (0.3,85.05) (0.35,85.29) (0.4,85.29) (0.45,85.25)};
            \addplot[color=cyan, thick, dashed] coordinates {(0.2,77.17) (0.25,81.92) (0.3,83.32) (0.35,83.41) (0.4,82.77) (0.45,81.27)};
            \addplot[color=cyan, thick] coordinates {(0.2,83.01) (0.25,85.54) (0.3,85.90) (0.35,85.36) (0.4,84.67) (0.45,83.09)};
            \end{axis}
        \end{tikzpicture}
    \end{minipage}
    \caption{Ablation study results showing the sensitivity of Point-GN's performance to key hyperparameters: (a) Number of neighbors (\(K\)), (b) Dimension of Gaussian Positional Encoding (GPE), (c) Number of stages, and (d) Sigma (\(\sigma\)). We compare the performance of the model on ModelNet40 \cite{wu20153d} (orange) and ScanObjectNN \cite{uy2019revisiting} (cyan) datasets, showing both the average and best performances.}

    \label{fig:ablation_study}
\end{figure*}
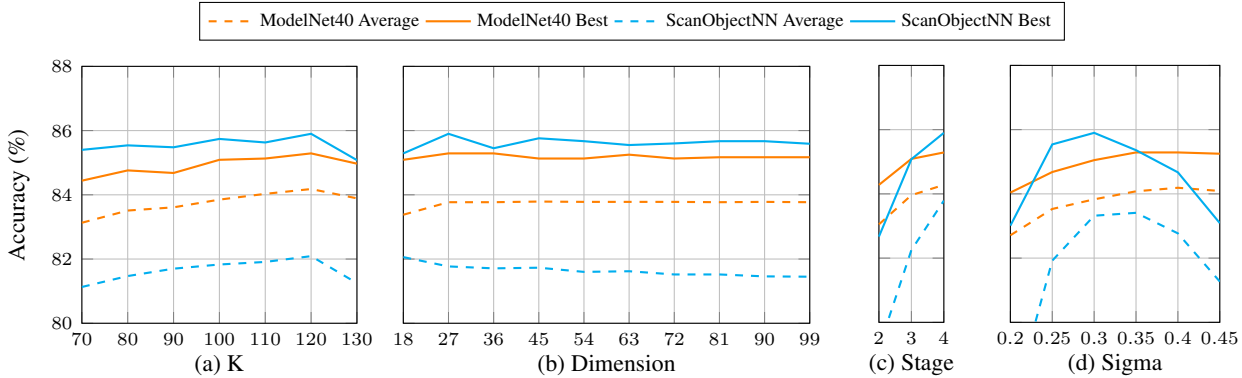

\subsection{Few-shot Classification on ModelNet40}

In the few-shot classification task on ModelNet40 \cite{wu20153d} (\cref{tab:fewshot}), Point-GN outperforms existing methods, demonstrating the best performance in both the 5-way and 10-way settings. Notably, non-parametric methods, including Point-NN \cite{zhang2023parameter} and Point-GN, significantly outperform parametric models in this scenario. While traditional deep learning models with learnable parameters often struggle with overfitting when only a small number of training samples are available, both Point-NN and Point-GN manage to avoid this pitfall, achieving superior generalization.

Point-GN, with its non-parametric design, achieves higher accuracy than Point-NN \cite{zhang2023parameter} in 3 out of 4 configurations (both the 5-way and 10-way 10-shot and 20-shot settings). In the remaining configuration (5-way 20-shot), Point-GN matches the performance of Point-NN, underscoring the consistency and robustness of our approach. Compared to traditional parametric methods, such as PointNet \cite{qi2017pointnet} and PointNet++ \cite{qi2017pointnet++}, which require more complex training and a larger number of parameters, Point-GN excels despite having zero trainable parameters, proving its effectiveness in few-shot learning scenarios.

These results highlight the power of our non-parametric method, which can achieve high accuracy with limited data and fewer resources, making it particularly suitable for applications with constrained training data.

\subsection{Computational Complexity Analysis}

We evaluate the computational efficiency of Point-GN in \cref{fig:test_speed}, where we compare its inference speed to Point-NN \cite{zhang2023parameter} by running both models on our system. Despite having zero trainable parameters, Point-GN demonstrates significantly faster inference across both the ScanObjectNN \cite{uy2019revisiting} and ModelNet40 \cite{wu20153d} datasets. This efficiency makes Point-GN ideal for real-time applications, such as autonomous driving and robotic perception.

In contrast, Point-NN \cite{zhang2023parameter}, while also non-parametric, achieves slower inference speeds, highlighting Point-GN as the more suitable choice for time-sensitive tasks that require both speed and high performance.

\subsection{Ablation Study}

We conducted an ablation study to evaluate the impact of different model configurations on the performance of Point-GN. Specifically, we examined the effect of four key factors: the number of neighbors \( k \) in the k-Nearest Neighbors algorithm, the dimension of Gaussian Positional Encoding (GPE), the number of stages in the model and the standard deviation \( \sigma \) of the Gaussian function.

\textbf{Effect of Number of Neighbors (\(k\)).} The number of neighbors \( k \) used in the k-NN algorithm also plays a critical role in model performance. Our experiments show that \textbf{\(k = \textbf{120} \)} offers the best trade-off between computational efficiency and accuracy. Smaller values, such as \( k = 70 \), fail to capture sufficient local context, while larger values, like \( k = 130 \), introduce irrelevant neighbors that confuse the model, hindering its ability to discern fine-grained geometric features. The performance comparison is shown in \cref{fig:ablation_study} (a).


\textbf{Effect of GPE Dimension.} The dimensionality of the Gaussian Positional Encoding (GPE) significantly affects model performance. As shown in \cref{fig:ablation_study} (b), increasing the dimension up to \textbf{27} improves accuracy, achieving the best performance in both ModelNet40 and ScanObjectNN datasets. However, beyond \textbf{45}, further increases in dimension result in diminishing returns, with accuracy slightly declining. This suggests that while higher dimensions can capture more complex features, excessively high dimensions add unnecessary complexity without substantial performance gains.

\textbf{Effect of Number of Stages.} Increasing the number of stages in the model generally improves accuracy. The \textbf{4-stage configuration} achieves the highest accuracy of \textbf{85.3\%} on the ModelNet40 dataset and \textbf{85.9\%} on the ScanObjectNN dataset, suggesting that deeper models are better equipped to capture complex spatial relationships and improve classification performance. Refer to \cref{fig:ablation_study} (c) for a detailed comparison of performance across different stage configurations.


\textbf{Impact of Sigma (\(\sigma\)).} The \(\sigma\) 
parameter of the Gaussian kernel (\(\sigma\)) determines the degree of locality in feature aggregation. For the ModelNet40 dataset, \(\sigma = \textbf{0.35}\) and \(\sigma = \textbf{0.4}\) yield the highest accuracy, effectively preserving both local and broader spatial contexts. On the other hand, for the ScanObjectNN dataset, we find that \(\sigma = \textbf{0.3}\) is the most effective, achieving the best accuracy by maintaining a balance between capturing fine-grained details and minimizing noise.  Smaller values (\(\sigma < 0.3\)) fail to aggregate sufficient local features, while larger values (\(\sigma > 0.4\)) introduce excessive smoothing, negatively impacting performance. These trends are illustrated in \cref{fig:ablation_study} (d).

\section{Conclusion}

In this paper, we introduced Point-GN a novel non-parametric network for 3D point cloud classification that combines Gaussian Positional Encoding (GPE) with non-learnable components such as FPS and KNN to efficiently capture both local and global geometric structures. By eliminating the need for learnable parameters, Point-GN provides a highly efficient and lightweight model suitable for real-time and resource-constrained environments. Experimental results on ModelNet40 and ScanObjectNN demonstrate that Point-GN achieves competitive accuracy, outperforming existing non-parametric methods while requiring zero trainable parameters and delivering fast inference speeds. 

For future work, we plan to extend the non-parametric framework of Point-GN by incorporating additional point cloud features and exploring its potential in more complex 3D tasks, such as semantic segmentation and object detection, further enhancing the model's versatility and real-world applicability.

{\small
\bibliographystyle{ieee_fullname}
\bibliography{egbib}
}

\end{document}